# Estimating Structural Disparities for Face Models

Shervin Ardeshir
Netflix
Los Gatos, CA
shervina@netflix.com

Cristina Segalin
Netflix
Los Gatos, CA
csegalin@netflix.com

Nathan Kallus
Cornell University & Netflix
New York, NY
kallus@cornell.edu

## Abstract

*In machine learning, disparity metrics are often defined by measuring the difference in the performance or outcome of a model, across different sub-populations (groups) of datapoints. Thus, the inputs to disparity quantification consist of a model's predictions $\hat{y}$, the ground-truth labels for the predictions $y$, and group labels $g$ for the data points. Performance of the model for each group is calculated by comparing $\hat{y}$ and $y$ for the datapoints within a specific group, and as a result, disparity of performance across the different groups can be calculated. In many real world scenarios however, group labels ($g$) may not be available at scale during training and validation time, or collecting them might not be feasible or desirable as they could often be sensitive information. As a result, evaluating disparity metrics across categorical groups would not be feasible. On the other hand, in many scenarios noisy groupings may be obtainable using some form of a proxy, which would allow measuring disparity metrics across sub-populations. Here we explore performing such analysis on computer vision models trained on human faces, and on tasks such as face attribute prediction and affect estimation. Our experiments indicate that embeddings resulting from an off-the-shelf face recognition model, could meaningfully serve as a proxy for such estimation.*

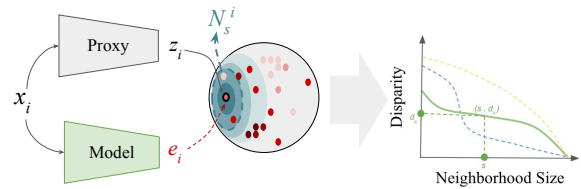

Figure 1. Overview of the proposed approach. Using the proxy (face recognition) embeddings, a neighborhood, and thus a neighborhood error for each datapoint is calculated. We show that the disparity in the performance of a model across different neighborhoods, could roughly estimate it's performance dispriy across individuals (for small neighborhood sizes), and across groups (for larger neighborhood sizes). In the toy example illustrated here, analysis on the model results in the green curve. Comparing that to other hypothetical baselines (blue and yellow), the green model is estimated to have the lowest performance disparity across individuals, as it has lower disparity for smaller neighborhood sizes. However, the blue curve is estimated to have lower group-level disparity, as it has lower estimated disparity in large neighborhood sizes. In this work, we empirically verify this phenomenon for face models, and when face recognition embeddings are used as proxy features.

## 1. Introduction

Studying bias in machine learning has recently become an active area of research [3, 27]. Inconsistent performance of machine learning models on sensitive data could have dire consequences, if/when the technology gets adopted in areas such as healthcare [5], financial loan applications [2], law enforcement [31], etc. As a result, measuring the performance of machine learning models across different groups, and ensuring that the models are not learning shortcuts based on irrelevant demographic-related confounding variables would be crucial. One of the sensitive types of data often used in training machine learning models, is images of human faces. A face image could reveal demographic-based information about individuals, from which the model's bias should be mitigated. Also, there is the known issue of data imbalance in many face-based data sets [19], which could lead to models learning shortcuts and therefore, ignoring less represented groups in the dataset.

There has been research conducted in the area of disparity of performance of face-model such as face detection [1], face recognition [19], generation [38], expression prediction [39], etc, focused on using demographic information for directly measuring, and often for training fairer models.

In many scenarios however, such data is not available for evaluation, and collecting it may be not scalable and/or desirable. In this work, our goal is to empirically verify whether we can estimate performance disparity of a face model, under such constraints, i.e. without having the demographic information. We show that feature embeddings extracted from an off-the-shelf face-recognition model could act as a proxy for estimating the performance disparity of an arbitrary face-model. We show that without any categorical information, and using proxy feature embedding from face recognition models, we can estimate disparity measures, which consistently correlate with both individual, and also group-level disparity metrics.

Prior work has also explored creating balanced datasets such as [17]. While being extremely valuable sources, scaling the number of samples in such datasets is often strenuous, as they require significantly higher per-datapoint annotation cost. In addition, the concept of a *fully balanced* dataset is non-trivial, and is often defined based on a limited number of pre-defined semantic attributes. Thus we believe that being able to measure performance disparity in a data-driven way, and on (larger) datasets that lack such annotation and balance, could provide additional insight.

**Intuition:** Face-recognition models are often trained using metric learning objectives. The model is trained to learn an embedding space, in which every positive pair of faces are mapped close to each other, while every negative pair of faces will be pushed apart from each other. Given enough training data, the models learn to match faces by comparing less mutable features of a face, i.e. facial features. In other words, given a face, a nearest neighbor retrieval in the face recognition feature space would ideally retrieve all the face images of the same identity first, and then other identities with similar facial features. As a result, the neighborhood of a face in this space, could be considered as a *lookalike sub-population* that shares similar facial features. An example of such manifold is shown in Figure 2 for the test set of the celebA dataset [26]. Our study shows that one can measure the disparity of the performance of a face-model of interest (such as affect recognition, or attribute prediction), across different neighborhoods across this embedding-space, and infer a notion of performance disparity for the model. This simple approach yields to proxy-based fairness metrics which highly correlate with both individual, and group level fairness metrics given different definitions of embedding-space neighborhoods.

*Fairness under unawareness* has recently become an active area of research [6, 12] as mitigating bias without requiring label information is a prevalent scenario. Most earlier work in this area have been focused on optimization approaches such as invariant risk minimization [7], and distributionally robust optimization [14], in which worst case risk is minimized to mitigate bias. Related to the same line of work, [24] uses adversarial re-weighting to mitigate bias in training models. Our work falls under the category of *assessment* of fairness without demographics and using proxy features, similar to [6]. The difference between our work and the earlier works in this area is due to the nature of the proxy features. Given that our proxies are extracted from a model with a metric learning objective, they do not directly provide psuedo-labels for datapoint (likelihood of belonghing to any specific group). Rather, the structure of datapoints could implicitly be indicative of an underlying groupings. Due to that, rather than disparity across semantic categories, we aim to mesaure *structural* disparities in the proxy embedding space.

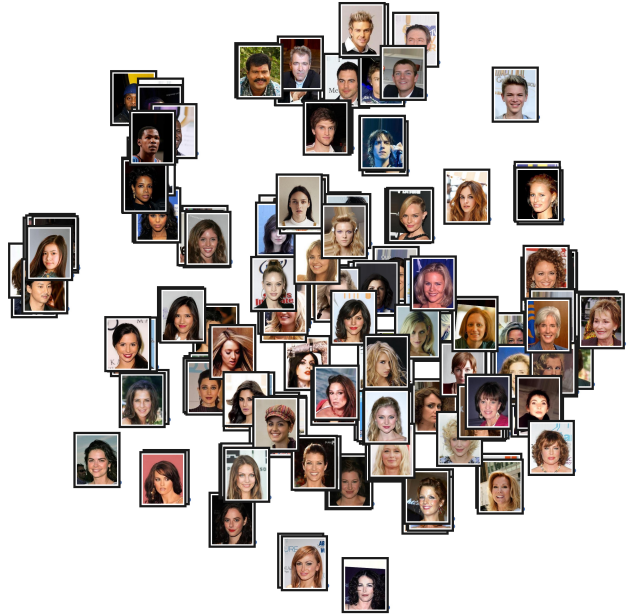

Figure 2. tSNE representation [36] of the face recognition embeddings on a random subset of the celebA dataset test set. We hypothesize that measuring disparity between the performance of a face model across different neighborhoods of this distribution is an informative measure, and predictive of its performance disparity metric across categories.

We evaluate performance disparity metrics for CNN models trained for face attribute prediction [26], and visual affect recognition [8, 13, 21, 23, 25, 35, 40], which aims to estimate the emotional state of a person using 2 continuous values of valence (how positive or negative the emotional display is), and arousal/intensity (how calming or exciting the emotional display is) [11, 29, 33].

## 2. Framework

Given a model trained to perform a task (such as predicting attributes or affect) on an input $x_i$ (face images), resulting in prediction $\hat{y}_i$, its performance/error could be evaluated by comparing $\hat{y}_i$ to its true value $y_i$. Given a group label $g_i$ for each data-point, performance disparity of the model could be evaluated by measuring the inconsistency of its performance across different groups. There-fore, given $n$ datapoints $\mathbf{X} = \{x_1, x_2, ..., x_n\}$, their corresponding model predictions $\hat{\mathbf{Y}} = \{\hat{y_1}, \hat{y_2}, ..., \hat{y_n}\}$, their true labels $\mathbf{Y} = \{y_1, y_2, ..., y_n\}$, and group memberships $\mathbf{G} = \{g_1, g_2, ..., g_n\}$ (in which each datapoint belongs to a group $g_i \in \{1, 2, ..., m\}$), the disparity of the model performance could be defined as a function of the three aforementioned inputs: $d = f(\mathbf{Y}, \hat{\mathbf{Y}}, \mathbf{G})$. Under unawareness, $\mathbf{G}$ is not available. Instead, a form of proxy feature $\mathbf{z}$ is available, which is assumed to implicitly contain information correlating with $\mathbf{G}$. The goal here is to calculate an approximation of the disparity metric: $\hat{d} = f'(\mathbf{Y}, \hat{\mathbf{Y}}, \mathbf{z})$ such that $d \sim \hat{d}$. In this work we do not use demographic labels $\mathbf{G}$ (individual and/or group labels) in training or validation, but we use them in order to evaluate our hypothesis, namely $d \sim \hat{d}$. The natural question would be, what does $d \sim \hat{d}$ entail. Given that in most model training scenarios, many variations of a model are being trained using different architectures, loss functions, data augmentations, etc, being able to compare different models and rank them in terms of disparity could be of more utility to the community. Therefore, in addition to evaluating $|d - \hat{d}|$, we also evaluate whether the ranking resulting from $\hat{d}$ would mimic that of $d$.

### 2.1. Proxy features

We use feature embeddings from a face recognition model [18] as proxy features for identities. Assuming face recognition models generalize, the embedding space manifold of faces should be highly structured as a function of facial features. Neighboring faces are more likely to belong to the same individual, or different individuals with similar facial features. Therefore, neighboring faces are more likely to share identity-related group memberships. On that basis, if a model performs more uniformly across different neighborhoods in this embedding space, it is more likely to be fair across different individuals, and also across different groups.

### 2.2. Constructing Embedding Neighborhoods

We define the neighborhood of a datapoint as the set of datapoints in its vicinity in the proxy embedding space. We use notation $N_s^i$, referring to neighborhood with size $s$ of the $i^{th}$ datapoint. We define the size of a neighborhood in two different ways:

**Based on radius:** For the $i^{th}$ datapoint (face image $x_i$, and its corresponding proxy $z_i$), we define $N_{s=r}^i$ as its neighborhood with radius $r$. The neighborhood is defined as the set of all datapoints whose proxy is within distance $r$ of $z_i$ (as shown in Figure 3 - left). Formally:

$$N_{s=r}^i = \{x_j | \, |z_j - z_i| < r\}. \quad (1)$$

**Based on number of nearest-neighbors (kNN):** An alternative way of defining a neighborhood, would be based on the number of nearest neighbors (k in kNN, as shown in Figure 3 - right), denoted as $N_{s=k}^i$.

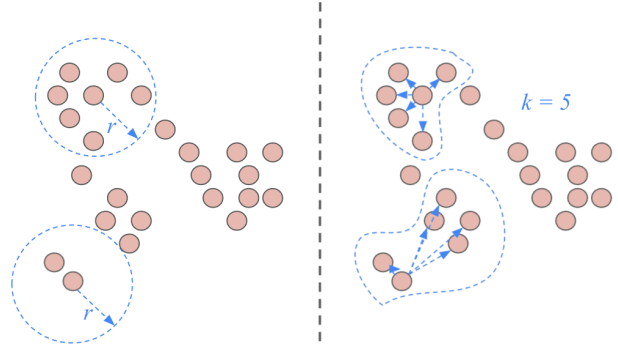

Figure 3. Left shows neighborhoods defined based on radius (distance) in the proxy space, resulting in a spatially uniform neighborhood size. Right shows the grouping based on k-nearest neighbors, resulting in neighborhoods with equal cardinality.

As shown in Figure 3, *radius*-based filtering would result in a spatially-balanced representation of the embedding space manifold. However, the cardinality of the neighborhood sets are going to be non-balanced, as different datapoints could have different number of neighbors within a fixed radius $r$. *kNN*-based filtering on the other hand, results in neighborhoods with balanced cardinality.

### 2.3. Disparity across Embedding Neighborhoods

Given the size $s$ neighborhood of the $i^{th}$ datapoint $N_s^i$, model's prediction error in that neighborhood could be quantified as:

$$e_s^i = \Sigma_{j \in N_s^i} \frac{|\hat{y}_j - y_j|}{|N_s^i|} \quad (2)$$

Given the performance of the model across different neighborhoods with the same size ($s = r$ or $s = k$), we can construct sets of neighborhood errors: $E_s = \{e_s^1, e_s^2, ..., e_s^N\}$, and measure the amount of disparity across them. We measure two standard notions of disparity, namely standard-deviation, and Rawlsian max-min [32]. Rawlsian max-min disparity is defined based on the ratio of minimum and maximum expected utility across groups, quantified as $1 - \frac{\min_i(e_i)}{\max_i(e_i)}|_{i \in G}$. Given that we are considering neighborhood errors to be surrogates for group errors, the Rawlsian max-min measure across neighborhoods of size $s$ would be defined as:

$$R_s = 1 - \frac{\min\limits_i e_s^i}{\max\limits_i e_s^i + \epsilon} \quad (3)$$

In which $\max\limits_i e_s^i$ would specify the error of the model on the most under-served neighborhood of size $s$. We also add a small constant value $\epsilon$ to simply prevent zero denominators. In addition to $R_s$, we also measure standard deviation $\sigma_s$ of error across different groups as an alternative measure of disparity. Given that these measures are based on embedding neighborhoods, we refer to them as DEN, short for **D**isparity across **E**mbedding **N**eighborhoods.

## 3. Experimental Results

### 3.1. Models and Datasets

We evaluate the efficacy of our disparity estimation method on a classification instance (attribute prediction), in addition to a regression instance (affect prediction). In the following we provide details on the setup for each instance.

#### 3.1.1 Attribute Prediction

We use celebA dataset [26] containing 40 attributes per image. We use the attributes *Male*, *Pale Skin*, and *Young* (which are often considered sensitive attributes), in addition to *Oval Face*, *Big Lips*, *Big Nose*, *Pointy Nose*, and *Narrow Eyes* (as they describe facial features), as different group labels. We then train a model to predict the other 32 attributes. We train a ResNet18 [15] model, which results in average attribute prediction accuracy of 83 %. In addition to the group labels, the dataset also provides identity labels for each of the faces, allowing us to validate the consistency of our approach with individual fairness, in addition to the different group fairness measures. We evaluate the 32 attribute predictors in terms of their performance disparity across neighborhoods, and compare their estimated performance disparity (DEN) to their actual ground-truth individual, and group-level disparity measures.

#### 3.1.2 Affect Recognition

We evaluate our approach on a regression instance of predicting valence and arousal. We train multiple models on the same tasks (predicting valence and arousal), and use the model prediction error to evaluate our approximate disparity metrics. We use the AFEW-VA [22] dataset alongside it's annotations of identity (individuals) and group (Male/Female shown as M/F), using which we measure the ground-truth individual and group-level performance disparities. The dataset consists of 600 videos extracted from movies, ranging from 10 to 120 frames. In our evaluations, faces are extracted from the videos and treated as single datapoints. We compare our estimated disparity measure (DEN) to the true individual and group-level disparity metrics. We use 6 different models, trained on the AffectNet dataset [28]. In the following, we provide details on the models used in our experiments.
**ResNet-18-IN** [15]: pretrained on Imagenet [9], fine-tuned for affect prediction.
**VGGFace** [30]: pretrained on VGGFace dataset [30], fine-tuned according to [20]-Table2.
**SE-ResNet-50** [4, 16]: pretrained on VGGFace2 [4], fine-tuned for affect prediction.
**DAN+** [37]: pretrained on VGGFace dataset [30], fine-tuned for affect prediction.
**EfficientNet-B4** [34]: pretrained on Imagenet [9], fine-tuned for affect prediction.
**EmoFAN** [35]: released by the papers' authors, directly used for inference. The model was trained already on AffectNet following similar training configurations and data transformations used in the aforementioned models.
For completeness, the performances of the models on AFEW-VA are reported in table 5 of the appendix. We also provide detailed descriptions of the standard affect recognition evaluation metrics in Section 5.1 of the appendix.

### 3.2. Evaluation:

In section 3.2.1, we describe the construction and interpretation of the Disparity across Embedding Neighborhoods (DEN) curves. In section 3.2.2 we compare our DEN measures, to the true individual and group level disparity measures, calculated using the ground-truth group information. We then measure the capability of DEN in terms of ranking different models in Section 3.2.3. Finally, in Section 3.2.4 we discuss the interplay between the proxy bias and the model bias.

#### 3.2.1 Disparity across Embedding Neighborhoods (DEN) Curves

In Figure 4, and 5 we illustrate the estimated disparity for each neighborhood radius (sweeping from minimum to maximum) for the celebA attribute predictors, and for the affect models respectively. Given each value for neighborhood size (such as radius $r$, or cardinality $k$), we measure the disparity measure of choice (Standard-deviation or Rawlsian max-min) of the model(s) performance across different neighborhoods with that neighborhood size. This results in a curve for each model (as shown in Figure 4 and 5). Y-axis shows the disparity across neighborhoods metric and x axis shows the neighborhood size. The values in the figure legend specify the area under each curve. Larger neighborhood sizes lead to more overlap between the neighborhoods, and therefore often less disparity across them. For the largest neighborhood size, all datapoints will be in the same neighborhood, resulting in disparity of 0 by definition. Also, in the smallest neighborhood size (1 datapoint per neighborhood), there is no structural information, resulting in disparity across all the datapoints. An ideally fair model would have a very small area under curve as it would perform with small disparity across different neighborhood sizes. The area under curve could potentially be seen as an overall measure of structural disparity based on facial features. Due to space limitations, we included standard-deviation based curves in the appendix.

### 3.2.2 Disparity Estimation Error

Given each neighborhood size ($s$), we measure the neighborhood errors for all of the datapoints. Disparity across neighborhood errors is then measured based on Rawlsian max-min $R_s$ ($R_r$ or $R_k$), as explained in Equation 3, or simply based on the standard deviation of the estimated errors, $\sigma_s$ ($\sigma_r$, or $\sigma_k$). We then evaluate the mean absolute error of different neighborhood-based disparity metrics and individual and group disparity metrics (Rawlsian: $R_G$, standard-deviation: $\sigma_G$). We quantify the mean absolute error between the estimated, and true disparity measure ($|R_s - R_G|$, and $|\sigma_s - \sigma_G|$). We evaluate the aforementioned absolute error for different neighborhood sizes, and different types of neighborhood (radius-based and kNN based). Figure 6 shows this error for the celebA dataset, and for the attribute prediction tasks (average error of 32 attributes). Top shows estimation of standard-deviation and bottom shows Rawlsian max-min disparity metric estimation error. For both metrics, left, shows radius based neighborhoods and right shows kNN based neighborhood definitions. In all the figures, it can be observed that smaller neighborhood sizes are better estimates for disparity-across-individuals, and larger neighborhood sizes, are better estimates of group-level-disparity. Similarly, Figure 7 shows the same patterns in the AFEW-VA dataset. In both experiments, we observe that individual disparity is better estimated with smaller neighborhoods, and group disparity, with larger neighborhoods. Please note that we do not prescribe using a specific neighborhood size (radius, or kNN), for any specific group-level disparity estimation, as they could be completely different based on the dataset, disparity metric, and proxy model. We simply conclude that smaller neighborhoods better approximate for individual disparity, and larger neighborhoods better approximate group-level disparities.

Please note that given no demographic information and no proxy, the only alternative measure of disparity would be calculating the disparity metrics across individual datapoints, or equivalently at $s = 0$. In all the curves in Figure 6 and 7, it can be observed that, a more optimal estimation error is achieved at a non-zero neighborhood sizes. This attests that using proxy neighborhoods is a better alternative when no group information is available.

### 3.2.3 Disparity-based Ranking

In most applications such as hyper-parameter search, architecture search, etc, being able to rank different models in terms of their performance disparity is often of more utility, compared to the absolute value of an estimated met-rics at a specific neighborhood size. Therefore, we evaluate how well the area under curve of our neighborhood-based metric (AUC-DEN) can rank different models in terms of their performance disparity. More specifically, given a set of models, we evaluate the area under the DEN curve. In addition we measure the disparity of the models across individuals and groups. We then evaluate if our ranking is consistent with the ranking acquired by the true individual and group level metrics. Table 1 and 2 contain the correlations for celebA and AFEW-VA datasets, respectively. We measure the Kendall-tau correlation coefficient between the two ranking lists, which evaluates the ratio of concordant pairs, resulting in a value between -1 and 1. It can be observed that for all of the different variations of neighborhood type (kNN or radius), disparity metrics (Rawlsian and standard-deviation), and across both datasets (celebA and AFEW-VA), there is a positive correlation (corr) with a statistically significant p-value. This shows consistent correlation between the estimated disparities and the true individual and group-level disparities in terms of ranking different models.

|  | kNN | | radius | |
|---|---|---|---|---|
|  | corr | p-val | corr | p-val |
| individual | 0.71 | 1.98e-10 | 0.7 | 4.97e-10 |
| Male | 0.54 | 4.22e-06 | 0.54 | 5.08e-06 |
| Young | 0.31 | 1.23e-02 | 0.3 | 1.63e-02 |
| Oval Face | 0.48 | 5.63e-05 | 0.5 | 2.49e-05 |
| Pale Skin | 0.41 | 8.46e-04 | 0.43 | 4.38e-04 |
| Big Lips | 0.28 | 2.54e-02 | 0.27 | 3.28e-02 |
| Big Nose | 0.3 | 1.48e-02 | 0.29 | 1.95e-02 |
| Pointy Nose | 0.46 | 1.64e-04 | 0.45 | 1.89e-04 |
| Narrow Eyes | 0.65 | 1.39e-08 | 0.66 | 6.34e-09 |

Table 1. The correlation between DEN-AUC (Area-under-curve of the Disparity across Embedding Neighborhoods), and the true disparity across individuals and groups in the celebA dataset for different definitions of neighborhood (radius and kNN). In all scenarios, the Kendall-tau correlation coefficient (corr) indicates a positive correlation with a statistically significant p-value, attesting that our estimated disparity metric is capable of meaningfully ranking different models in terms their true cross-group performance disparities.

|  | kNN | | radius | |
|---|---|---|---|---|
|  | corr | p-val | corr | p-val |
| individual | 0.79 | 1.07e-04 | 0.82 | 4.41e-05 |
| M/F | 0.42 | 6.29e-02 | 0.39 | 8.63e-02 |

Table 2. The correlation between DEN-AUC and group-disparity for the afewVA dataset. Similar to the celebA results shown in Table 1, we observe a consistent positive correlation between the rankings resulting from the estimated and true disparities, with statistically significant p-values.

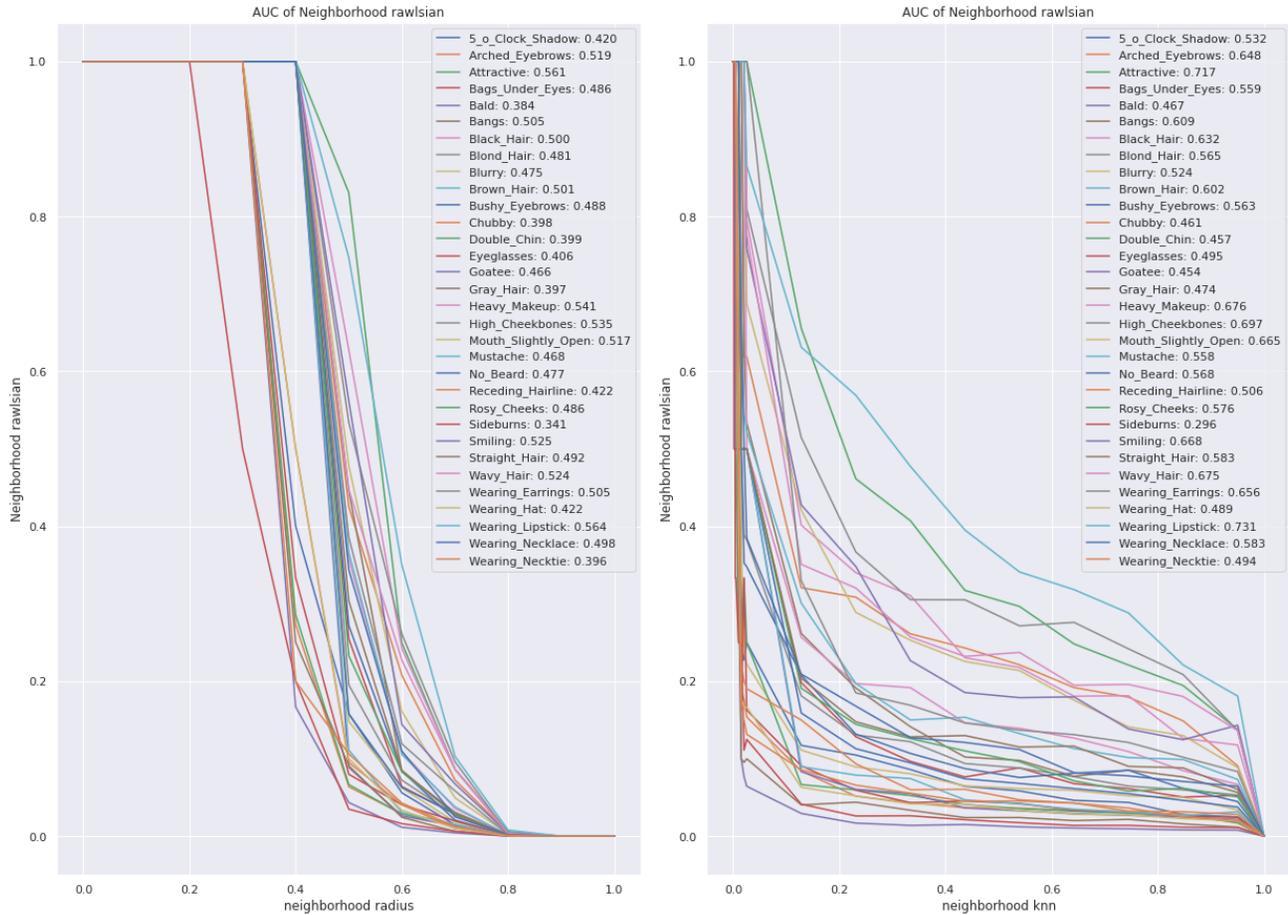

Figure 4. The Rawlsian Disparity across Embedding Neighborhoods (DEN) evaluated across different radiuses for different attribute predictors of the celebA dataset. The figure legend contains the area under curve for each attribute predictor. The attributes Attractive and Wearing Lipstick have been predicted to have the highest disparity.

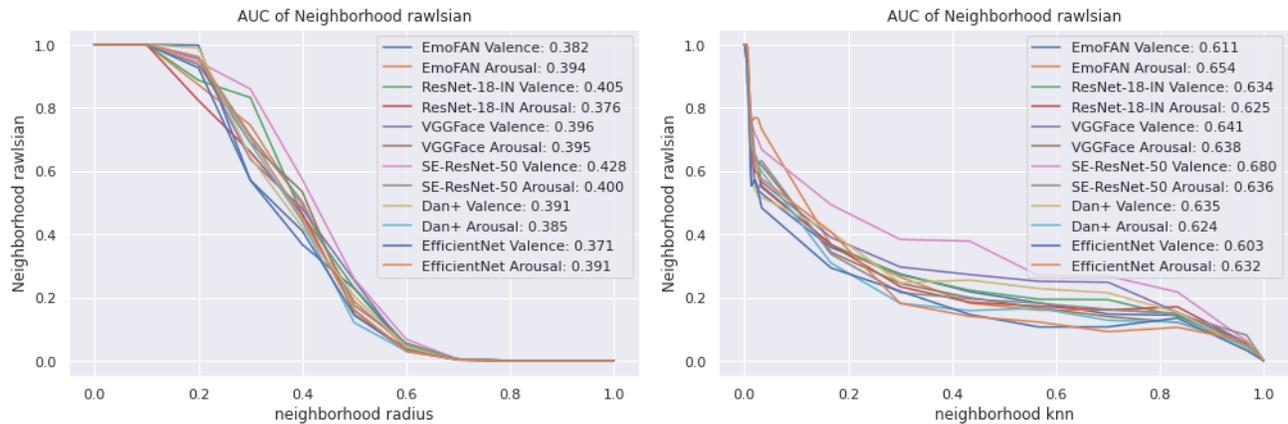

Figure 5. The Rawlsian Disparity across Embedding Neighborhoods (DEN) evaluated across different radiuses for different models predicting Valence and Arousal. Comparing EmoFAN Valence (blue) to ResNet-18 Valence (green), it can be observed that ResNet18 has lower disparity for smaller radiuses (thus more likely to be a better choice in terms of individual fairness), and EmoFAN seems to have lower disparity at larger radiuses, thus, more likely to be fairer across groups. Given this pool of models, one may pick valence predictor of EfficientNet (lowest AUC across valence predictors), and ResNet-18-IN (or DAN+) for Arousal prediction (lowest AUC across arousal predictors).

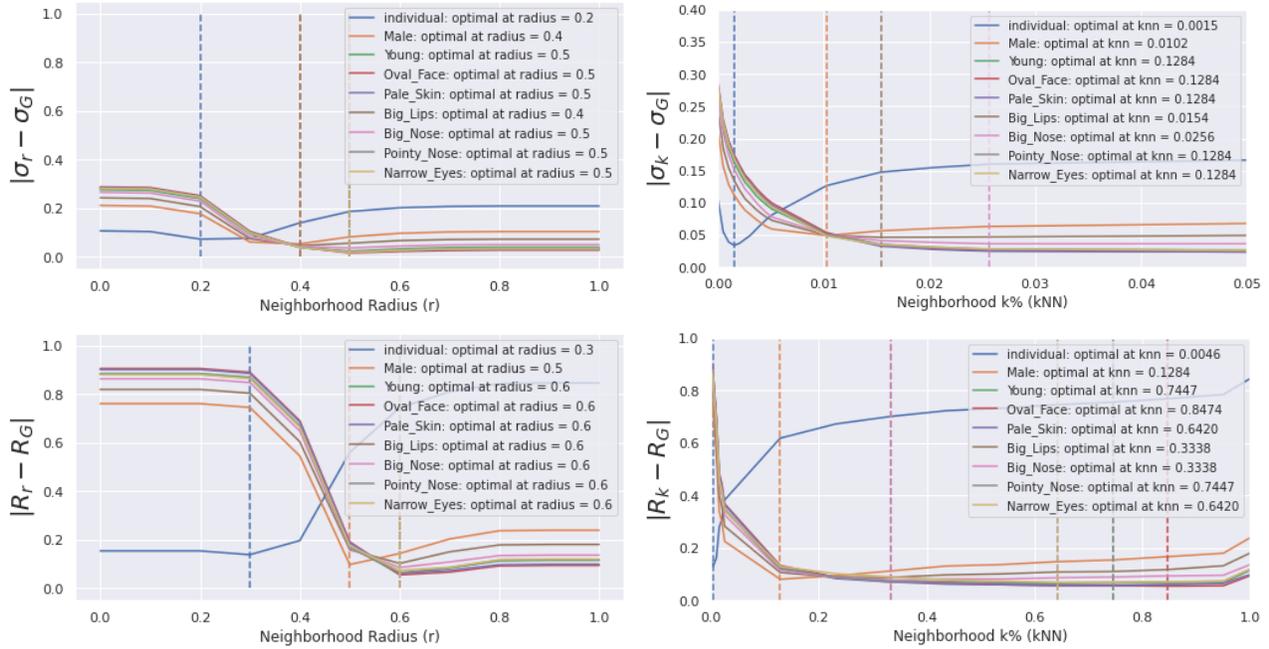

Figure 6. Disparity estimation error for different neighborhood sizes for the celebA dataset. Top-row shows the error of estimating the standard deviation across individuals. Bottom row shows the Rawlsian max-min disparity estimation error. We evaluate different neighborhood sizes based on radius (left), and kNN (right). In all variations, minimum disparity estimation error for individuals consistently occurs in smaller neighborhood sizes compared to group-level disparity measures.

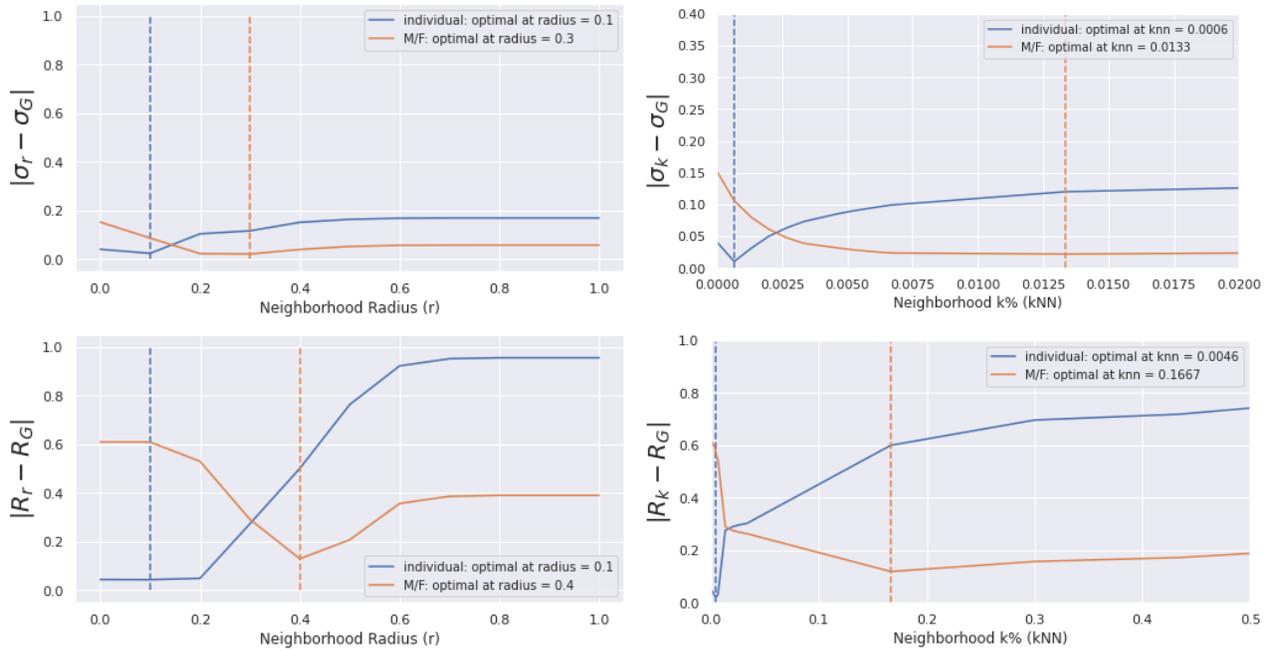

Figure 7. Disparity estimation error for different neighborhood sizes for the AFEW-VA dataset. Top-row shows the error of estimating the standard deviation across individuals. Bottom row shows the Rawlsian max-min disparity estimation error. We evaluate different neighborhood sizes based on radius (left), and kNN (right). In all variations, minimum disparity estimation error for individuals consistntly occurs in smaller neighborhood sizes compared to group-level disparity metrics.

| Model | PCC | p-value |
|---|---|---|
| Resnet18 - Valence | 0.027 | 0.27 |
| Resnet18 - Arousal | 0.045 | 0.07 |
| VGGFace - Valence | -0.00 | 0.79 |
| VGGFace - Arousal | 0.040 | 0.11 |
| SE-ResNet-50 - Valence | 0.009 | 0.72 |
| SE-ResNet-50 - Arousal | 0.024 | 0.34 |
| DAN+ - Valence | 0.000 | 0.97 |
| DAN+ - Arousal | 0.019 | 0.43 |
| EfficientNet-B4 - Valence | -0.00 | 0.98 |
| EfficientNet-B4 - Arousal | 0.020 | 0.42 |
| EmoFAN - Valence | -0.01 | 0.67 |
| EmoFAN - Arousal | 0.016 | 0.51 |

Table 3. Pearson Correlation Coefficient of Proxy-noise vs model error AFEW-VA dataset.

### 3.2.4 Proxy-noise vs estimation error

Given that we use features extracted from a face recognition model as a proxy, and the fact that this model itself may have biased performance across different sub-populations of the dataset, the interplay of the two biases is non-trivial. To measure the independence of these biases, we measure the correlation between the performance of each model on each datapoint, and the retrieval metric of the same datapoint in the proxy space (retrieving faces of the same person in the dataset). The pearson correlation coefficient between per-datapoint attribute prediction accuracy for each attribute, and per-datapoint face-recognition AUROC is reported in Table 4 and 3 for celebA and AFEW-VA, respectively. It can be observed that there is no significant correlation in these instances. Even though we simply verified low correlation between the performance of the model and the proxy in our experiments, the interplay between the proxy inaccuracy and the model inaccuracy should be further investigated, as the effects in a non-correlated scenario is non-trivial.

## 4. Conclusion

This work shows the possibility of estimating disparity of face models, on datasets with no demographic information. This opens the door to more thorough evaluation of face-models on larger datasets, which mostly do not contain such information. Given that this analysis is done on proxies, it measures structural bias of models towards arbitrary facial features. Thus, even in presence of demographic information, it could provide complementary insight. We believe that this analysis would not be a replacement to group aware analysis, but merely a complement, and an insightful alternative in the absence of group information.

| Model | PCC | p-value |
|---|---|---|
| 5o Clock Shadow | -0.03 | 0.16 |
| Arched Eyebrows | -0.01 | 0.53 |
| Attractive | 0.003 | 0.87 |
| Bags Under Eyes | -0.03 | 0.13 |
| Bald | -0.02 | 0.29 |
| Bangs | 0.009 | 0.66 |
| Black Hair | -0.02 | 0.33 |
| Blond Hair | -0.00 | 0.77 |
| Blurry | 0.040 | 0.07 |
| Brown Hair | 0.027 | 0.22 |
| Bushy Eyebrows | -0.02 | 0.34 |
| Chubby | -0.02 | 0.26 |
| Double Chin | -0.02 | 0.36 |
| Eyeglasses | -0.01 | 0.60 |
| Goatee | -0.00 | 0.90 |
| Gray Hair | 0.001 | 0.96 |
| Heavy Makeup | 0.004 | 0.83 |
| High Cheekbones | -0.02 | 0.23 |
| Mouth Slightly Open | -0.01 | 0.38 |
| Mustache | -0.00 | 0.88 |
| No Beard | -0.03 | 0.13 |
| Receding Hairline | -0.00 | 0.97 |
| Rosy Cheeks | 0.002 | 0.90 |
| Sideburns | -0.01 | 0.45 |
| Smiling | -0.01 | 0.46 |
| Straight Hair | 0.007 | 0.72 |
| Wavy Hair | -0.00 | 0.76 |
| Wearing Earrings | -0.00 | 0.78 |
| Wearing Hat | 0.014 | 0.51 |
| Wearing Lipstick | 0.036 | 0.10 |
| Wearing Necklace | 0.011 | 0.60 |
| Wearing Necktie | -0.01 | 0.46 |

Table 4. Pearson Correlation Coefficient of Proxy-noise vs model error on Celeb-A dataset.

# 5. Appendix

Here, we provide details on training and evaluation of the affect recognition models used in Section 5.1, and more variations of the DEN curves in Section 5.2.

## 5.1. Performance of Affect Recognition Models

The performance metrics used during the evaluation of the Valence-Arousal models are: RMSE, CCC, PCC, and SAGR. Given $\hat{Y}$ as predicted label, $Y$ the ground-truth label, $(\mu_Y, \sigma_Y)$, and $(\mu_{\hat{Y}}, \sigma_{\hat{Y}})$ being their mean and standard deviation, the metrics are defined as follows.
Root mean square error (RMSE): evaluates how close the predicted values are from the ground-truth values:

$$RSME(Y, \hat{Y}) = \sqrt{\mathbb{E}((Y - \hat{Y})^2)}. \quad (4)$$

Sign agreement (SAGR) evaluates whether the sign of predicted and ground-truth values match:

$$SAGR(Y, \hat{Y}) = \frac{1}{n} \sum_{i=1}^{n} \delta(\text{sign}(y_i), \text{sign}(\hat{y}_i)) \quad (5)$$

Pearson correlation coefficient (PCC) evaluates how predictions and ground-truth values correlates:

$$PCC(Y, \hat{Y}) = \frac{\mathbb{E}(Y - \mu_Y)(\hat{Y} - \mu_{\hat{Y}})}{\sigma_Y \sigma_{\hat{Y}}} \quad (6)$$

Concordance correlation coefficient (CCC) incorporated the PCC value but penalizes correlated signals with different means:

$$CCC(Y, \hat{Y}) = \frac{2\sigma_Y \sigma_{\hat{Y}} PCC(Y, \hat{Y})}{\sigma_Y^2 + \sigma_{\hat{Y}}^2 + (\mu_y - \mu_{\hat{Y}})^2} \quad (7)$$

All models have been trained using learning rate of 0.0001, weight decay of 0, and Adam optimizer. Batch size was set to 128, and we decreased the learning rate every 15 epoch by a factor of 10. For the loss function, we combined a cross entropy loss $\mathcal{L}_{\text{categories}}$ for the discrete emotions and a sum of three regression loss terms (RMSE, CCC, PCC), regularized with shake-shake regularization coefficients following [10]. The parameters $\alpha$, $\beta$ and $\gamma$ randomly and uniformly sampled in the range of (0, 1). The regression loss is a combination of three terms: RMSE, CCC, PCC.

$\mathcal{L}(Y, \hat{Y}) = \mathcal{L}_{categories} + \frac{\alpha}{\alpha+\beta+\gamma} \mathcal{L}_{MSE}(Y, \hat{Y}) + \frac{\beta}{\alpha+\beta+\gamma} \mathcal{L}_{PCC}(Y, \hat{Y}) + \frac{\gamma}{\alpha+\beta+\gamma} \mathcal{L}_{CCC}(Y, \hat{Y})$
with
$\mathcal{L}_{\text{categories}} = -\sum_{i=1}^{n} \hat{y}_i \log(y_i)$
$\mathcal{L}_{MSE} = \text{MSE}_{\text{valence}}(Y, \hat{Y}) + \text{MSE}_{\text{arousal}}(Y, \hat{Y})$
$\mathcal{L}_{PCC} = 1 - \frac{\text{PCC}_{\text{valence}}(Y, \hat{Y}) + \text{PCC}_{\text{arousal}}(Y, \hat{Y})}{2}$
$\mathcal{L}_{CCC} = 1 - \frac{\text{CCC}_{\text{valence}}(Y, \hat{Y}) + \text{CCC}_{\text{arousal}}(Y, \hat{Y})}{2}$

## 5.2. DEN Curves

Figure 8 and 9 illustrate the DEN curves for different neighborhood types (knn and radius), and for different disparity measures (standard-deviation, and Rawlsian max-min)

| Model | RSME (V,A) | SAGR (V,A) | PCC (V,A) | CCC (V,A) |
|---|---|---|---|---|
| Resnet18 | 0.34, 0.35 | 0.62, 0.75 | 0.46, 0.43 | 0.42, 0.37 |
| VGGFace | 0.32, 0.35 | 0.60, 0.75 | 0.50, 0.41 | 0.47, 0.37 |
| SE-ResNet-50 | 0.43, 0.39 | 0.63, 0.74 | 0.49, 0.41 | 0.42, 0.34 |
| DAN+ | 0.35, 0.31 | 0.61, 0.75 | 0.46, 0.44 | 0.41, 0.42 |
| EfficientNet-B4 | 0.32, 0.32 | 0.59, 0.72 | 0.49, 0.44 | 0.47, 0.41 |
| EmoFAN | 0.28, 0.30 | 0.60, 0.77 | 0.44, 0.38 | 0.43, 0.34 |

Table 5. Performance metrics on the AFEW-VA dataset. (V,A) stands for (Valence, Arousal) using standard affect recognition evaluation metrics, as explained in 5.1.

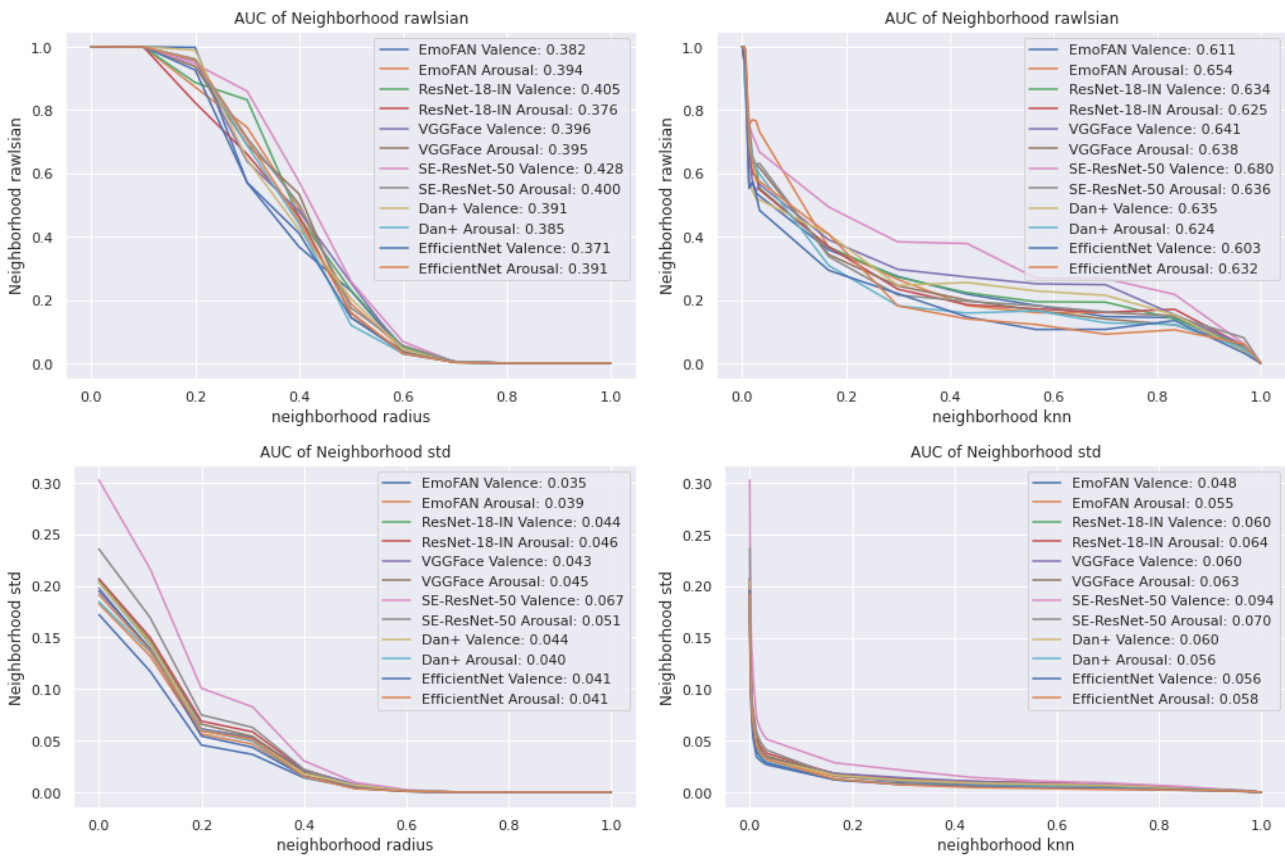

Figure 8. All variations of the DEN curves on the AFEW-VA dataset. Top shows Rawlsian max-min disparity metric, and the bottom row shows the standard-deviation disparity. Left shows radius based neighborhoods, and right shows knn based neighborhoods.

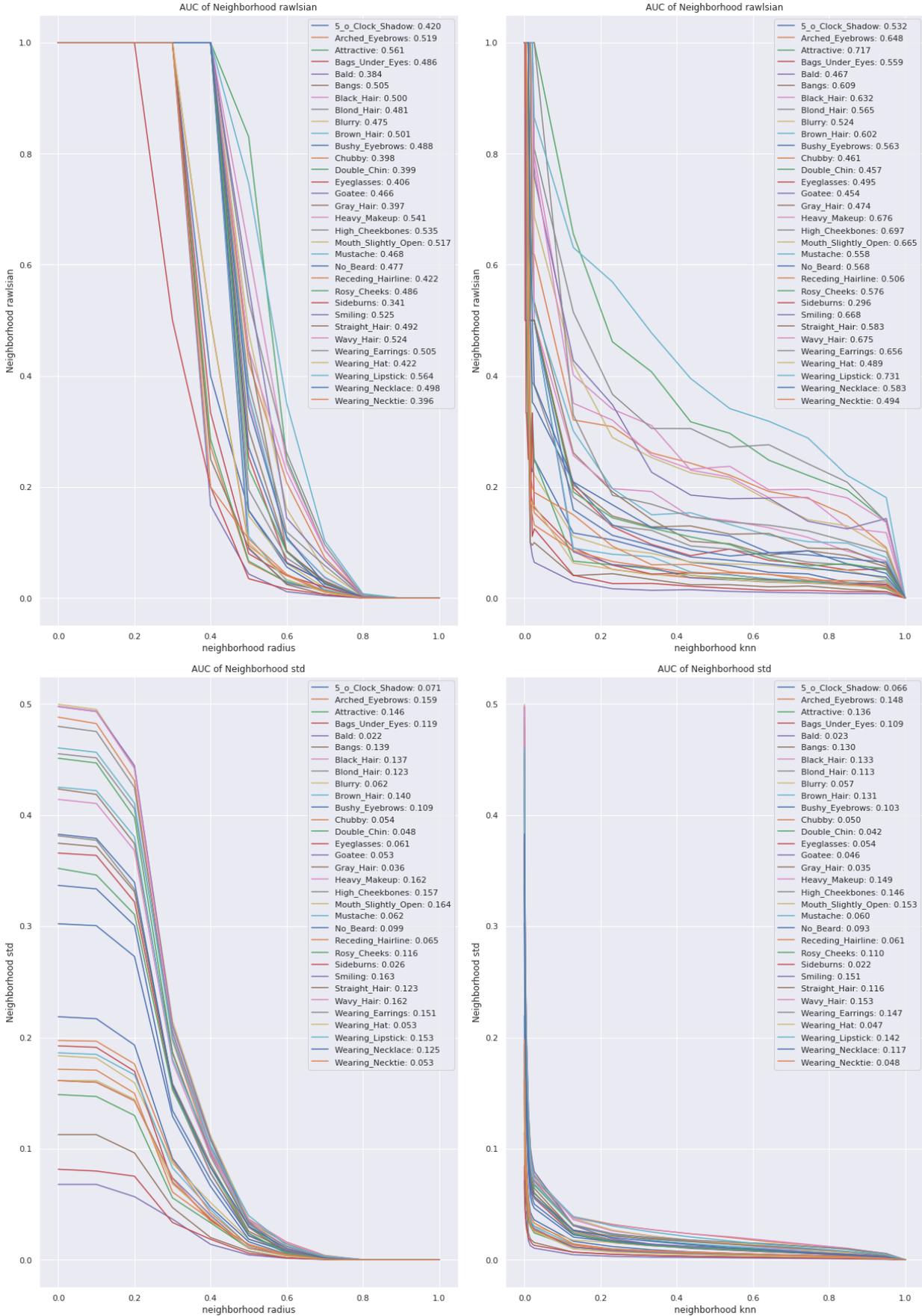

Figure 9. All variations of the DEN curves on the Celeb-A dataset. Top shows Rawlsian max-min disparity metric, and the bottom row shows the standard-deviation disparity. Left shows radius based neighborhoods, and right shows knn based neighborhoods.